\documentclass{article} 
\usepackage[T1]{fontenc}
\usepackage{arxiv,times}

\usepackage{amsmath,amsfonts,bm}

\def\eqref#1{equation~\ref{#1}}

\def\1{\bm{1}}

\DeclareMathAlphabet{\mathsfit}{\encodingdefault}{\sfdefault}{m}{sl}
\SetMathAlphabet{\mathsfit}{bold}{\encodingdefault}{\sfdefault}{bx}{n}

\usepackage{hyperref}
\usepackage{xurl}
\usepackage{graphicx}
\usepackage{booktabs}
\usepackage{multirow}
\usepackage{etoolbox}
\usepackage{caption}
\usepackage[table]{xcolor}
\usepackage[breakable,skins]{tcolorbox}

\newsavebox{\promptblockbox}
\newenvironment{promptblock}[1][Prompt]{%
    \par\smallskip\noindent
    \begin{lrbox}{\promptblockbox}%
    \begin{minipage}{\dimexpr\linewidth-2\fboxsep-2\fboxrule\relax}%
    \setlength{\parindent}{0pt}%
    \textbf{#1}\par\smallskip
    \ignorespaces
}{%
    \end{minipage}%
    \end{lrbox}%
    \fbox{\usebox{\promptblockbox}}%
    \par\smallskip
    \ignorespacesafterend
}

\definecolor{tideblue}{RGB}{232,242,252}
\DeclareCaptionFont{tableten}{\fontsize{10pt}{12pt}\selectfont}
\definecolor{tiretitleblue}{RGB}{60,99,155}
\definecolor{tiretitlegreen}{RGB}{39,84,23}
\title{\textcolor{tiretitleblue}{TI}\textcolor{tiretitlegreen}{RE}: \textcolor{tiretitleblue}{T}op-Down Formula \textcolor{tiretitleblue}{I}nstantiation \\with \textcolor{tiretitlegreen}{R}eliable \textcolor{tiretitlegreen}{E}xecution \\for Physics Problem Synthesis}

\author{%
Xiaoyu Wu\textsuperscript{1,*}\quad
Zhaoxin Yu\textsuperscript{2,3,*}\quad
Xinglin Xiao\textsuperscript{1}\quad
Nan Xu\textsuperscript{1,2,\ensuremath{\dagger}}\\[6pt]
\normalfont\small\textsuperscript{1}ScienceOne AI\\
\normalfont\small\textsuperscript{2}State Key Laboratory of Multimodal Artificial Intelligence Systems,\\
\normalfont\small Institute of Automation, Chinese Academy of Sciences\\
\normalfont\small\textsuperscript{3}School of Artificial Intelligence, University of Chinese Academy of Sciences\\[4pt]
\normalfont\small\textsuperscript{*}Equal contribution.\quad
\textsuperscript{\ensuremath{\dagger}}Corresponding author.\\
\normalfont\small\href{mailto:nan.xu@scienceone.ai}{\texttt{nan.xu@scienceone.ai}}
}
\date{}

\hypersetup{
    hidelinks,
    pdftitle={TIRE: Top-Down Formula Instantiation with Reliable Execution for Physics Problem Synthesis},
    pdfauthor={Xiaoyu Wu, Zhaoxin Yu, Xinglin Xiao, Nan Xu},
    pdfsubject={},
    pdfkeywords={}
}

\newcommand{\defaultteacher}{Qwen3-235B-A22B-Thinking-2507}

\begin{document}

\maketitle

\begin{abstract}
Advancing physics reasoning in large language models requires scalable training data that connect physical laws, concrete problem scenarios, and verifiable numerical answers.
Expert-designed problems offer top-down control over physical laws, conditions, and scenarios, but require costly human effort.
LLM-based synthesis from texts scales efficiently, but model-generated answers and assessments leave supervision vulnerable to hallucinations.
We introduce \textbf{TIRE}, a framework for Top-Down Formula Instantiation with Reliable Execution that brings knowledge-driven problem construction to large-scale data generation.
Top-down instantiation uses reusable physical formulas and their applicability conditions to control the physical relations and numerical configurations of generated problems.
Reliable execution establishes reference answers before problem generation, providing a common numerical foundation for problem construction and teacher distillation.
Together, these mechanisms combine control over problem content with execution-grounded supervision for both supervised fine-tuning and reinforcement learning.
Using TIRE, we construct \textbf{TIRE-19k}, spanning nine subfields of physics.
Across \textbf{3} backbone LLMs and \textbf{4} benchmarks, TIRE-19k outperforms \textbf{5} representative datasets with equal amounts of SFT data, exceeding the strongest baseline by \textbf{2.86 percentage points} on average.
Further GRPO raises Qwen3-8B's average accuracy by \textbf{1.52 percentage points} over its TIRE-19k SFT checkpoint.
These results demonstrate the value of generating physics training data from explicit physical knowledge with reliable execution.
\end{abstract}

\begin{center}
\begin{tcolorbox}[
    width=0.9\textwidth,
    colback=black!6,
    colframe=black!6,
    boxrule=0pt,
    arc=2.5mm,
    left=8pt, right=8pt, top=6pt, bottom=6pt,
    before skip=0pt, after skip=0pt,
    fontupper=\small
]
\textbf{Code:} \url{https://github.com/ScienceOne-AI/TIRE}\\[3pt]
\textbf{Dataset:} \url{https://huggingface.co/datasets/ScienceOne-AI/TIRE}
\end{tcolorbox}
\end{center}

\section{Introduction}
Physics problem solving requires models to translate physical scenarios described in natural language into appropriate physical models and derive verifiable answers through symbolic derivation and numerical computation.
It is therefore both an important task for evaluating the scientific reasoning capabilities of large language models (LLMs) and relevant to practical applications such as intelligent science education, engineering analysis, and research assistance \citep{SciBench, UGPhysics}.
The current limitations of models in physics reasoning are largely due to a shortage of high-quality, verifiable training data.
In mathematics, researchers have gradually established a systematic pipeline for generating verifiable data: they first construct executable programs, computational graphs, or formal proofs that encode valid solution processes, then convert them into natural-language problems and solutions, and finally use deterministic verifiers to independently and objectively check the correctness of the generated data \citep{MUSTARD, LeanWorkbook, DeepSeek-Prover-V2, RV-Syn}.
Compared with mathematical problem solving, in which derivations are typically carried out within a given formal structure, physics problem solving is more challenging.
It requires not only formal mathematical derivation and computation, but also consistency among real-world physical scenarios, applicable laws, their conditions of validity, and formal models \citep{SciBench, PhysicsReasoner, FormulaReasoning}.

Existing efforts to build physics problem-solving data broadly follow two routes.
One route relies on expert-designed problems from textbooks, examinations, and question banks \citep{TheoremQA, UGPhysics}.
Experts construct problems top-down, starting from physical laws and specifying conditions, parameters, and scenarios. This gives direct control over problem content and quality, but the required human effort limits dataset scale.
The other route uses LLMs to directly generate or complete problems, reasoning processes, and answers from existing texts \citep{SciInstruct, MEGAScience, PhysLogic}.
Although this approach scales efficiently, model hallucinations can introduce errors into problems and solutions, and quality assessment often relies on language models without an independent computational reference grounded in physics.
The physical laws and parameter values represented in the data also depend on source texts and model preferences, limiting control over problem construction.
The challenge is to combine top-down construction from physical knowledge with scalable generation and reliable supervision grounded in program execution.

Therefore, we propose TIRE, a framework for Top-Down Formula Instantiation with Reliable Execution that transforms structured physical knowledge into scalable training data for physics reasoning.
Top-down instantiation starts from reusable physics formulas and their applicability conditions, giving explicit control over the physical relations and numerical configurations underlying each problem.
TIRE translates these relations into validated executable programs, instantiates concrete parameter values, and constructs natural-language problems grounded in the resulting instances.
Reliable execution establishes reference answers before problem generation, providing a common numerical foundation for both problem construction and teacher distillation.
A teacher independently solves the generated problems, and its reasoning traces are selected by matching their final answers to these execution-derived references.
This integration of physical knowledge, controlled instantiation, and teacher distillation supports systematic data construction with reliable, execution-grounded numerical supervision for both supervised fine-tuning and reinforcement learning.
Using TIRE, we construct TIRE-19k, comprising 19,144 examples across nine subject areas and grounded in 3,051 formula records extracted from PHYSICS and SciBench.

Our main contributions are as follows:
\begin{enumerate}
    \item We propose TIRE, a framework that combines \emph{top-down formula instantiation} with \emph{reliable execution} to generate physics training data at scale. Physical knowledge guides problem construction, and deterministic program outputs provide reference answers for supervision.
    
    \item Based on TIRE, we construct the TIRE-19k dataset, covering nine subfields of physics with 19,144 examples grounded in 3,051 formula records. Each example includes an execution-derived reference answer and an answer-verified teacher reasoning trace.
    
    \item We conduct extensive experiments. Across three backbones and four benchmarks, TIRE-19k outperforms existing datasets in both in-domain performance and out-of-domain generalization with equal amounts of training data. Ablations support the framework's components, teacher comparisons demonstrate its effectiveness with multiple teachers, and GRPO yields further gains after SFT.
\end{enumerate}

\section{Related Work}
\paragraph{Physics Problem-Solving Datasets.}
Physics reasoning resources broadly follow two routes.
Expert curation supports carefully specified problems, but collecting and validating instances is costly to scale \citep{TheoremQA, SciBench, GPQA}.
LLM-based approaches expand data through question annotation, document-grounded synthesis, and reasoning distillation \citep{SciInstruct, PHYSICS, NaturalReasoning, MEGAScience, PhysLogic}.
Templates and topic hierarchies provide structure for generation \citep{QuantumLLMInstruct, SHARP}, while FormulaReasoning supplies formula annotations and a formula database, and filters calculation errors in problems augmented from question--explanation exemplars \citep{FormulaReasoning}.
The challenge is to combine scalable construction with explicit control over physical instances and reliable numerical supervision.
TIRE addresses this challenge through \emph{top-down} synthesis: reusable physical formulas and their applicability conditions guide numerical instantiation and subsequent natural-language problem construction.
Validated programs establish each instance's reference answer before question generation.
The same references then select independently generated teacher traces by final-answer agreement, linking controlled problem construction with execution-grounded supervision.

\paragraph{Program-Guided Verifiable Data Synthesis.}
RV-Syn composes reusable mathematical functions into executable graphs and back-translates them into problems, while AMD synthesizes programs from mathematical knowledge and tools; both use execution outputs to check generated answers \citep{RV-Syn, AMD}.
Formal methods use proof assistants to check statements and proofs when generating theorem-proving data \citep{MUSTARD, LeanWorkbook, DeepSeek-Prover-V2}; TP-as-a-Judge formalizes generated questions and answers and uses a prover to assess their reasoning steps \citep{TP-as-a-Judge}.
Physics additionally requires each numerical instance to correspond to a scenario in which the selected physical relation is applicable.
TIRE builds on the execution-guided paradigm by using formula records to guide parameter assignments and scenario construction.
Program interfaces specify physical meanings and units, and formula-derived tests check implementations before numerical instantiation.
Program outputs provide a shared numerical reference for problem generation and teacher-answer selection, connecting controlled physics instance construction with answer-verified supervision for SFT and reward references for GRPO.

\section{Method}
\label{sec:method}

\subsection{Overview}

\begin{figure}[t]
    \centering
    \includegraphics[width=\textwidth]{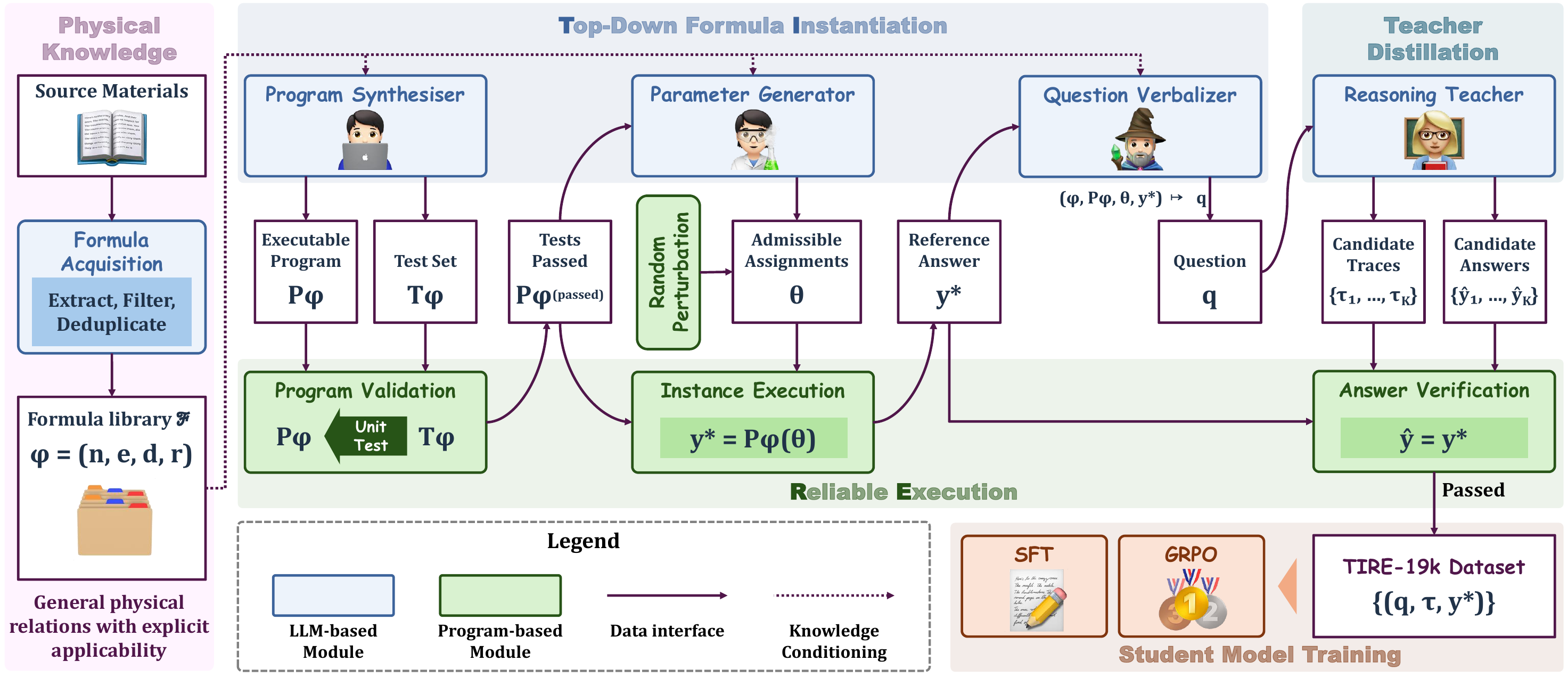}
    \caption{
    Overview of TIRE.
    Physical knowledge conditions formula instantiation, which produces questions for teacher distillation.
    Blue blocks denote LLM-based modules; green blocks denote program-based operations.
    Solid arrows indicate data transfer, and dotted arrows indicate knowledge conditioning.
    Execution-derived answers support both question construction and teacher-answer verification; retained examples support student training.
    }
    \label{fig:tide_framework}
\end{figure}

TIRE generates physics problem-solving training data in a top-down manner from the domain knowledge encoded in physics formulas.
Given a formula collection $\mathcal{F}$, it constructs a dataset $\mathcal{D}=\{(q_i,\tau_i,y_i^*)\}_{i=1}^{N}$, where $q_i$ is a natural-language problem, $y_i^*$ is an execution-derived reference answer, and $\tau_i$ is a teacher reasoning trace whose final answer agrees with $y_i^*$.
The framework comprises three modules: Physical Knowledge, Top-Down Formula Instantiation, and Teacher Distillation.
Physical Knowledge supplies reusable relations and their applicability conditions; formula instantiation turns these relations into numerical problems; teacher distillation supplies reasoning traces selected by numerical answer verification.
Program-based operations support reliable numerical supervision across these modules: Program Validation checks formula implementations, Instance Execution computes reference answers, and Answer Verification selects teacher traces.
LLMs make the semantic choices needed to instantiate physical relations and express problems, while program outputs provide a shared numerical reference across generation and selection.
The resulting data connect each natural-language problem to the formula-based numerical instance from which it was constructed.
We apply this framework to construct TIRE-19k, following the architecture in Figure~\ref{fig:tide_framework}.

\subsection{Physical Knowledge}

Formula Acquisition extracts general physical relations from source problems and their official solutions when available.\footnote{Sources are \href{https://github.com/yale-nlp/Physics}{PHYSICS} \citep{PHYSICS} and the physics subset of \href{https://huggingface.co/datasets/xw27/scibench}{SciBench} \citep{SciBench}. Extraction uses DeepSeek-R1-250528 \citep{DeepSeek-R1} (\href{https://huggingface.co/deepseek-ai/DeepSeek-R1-0528}{R1-0528 model card}).}
Each formula is represented as
\begin{equation}
    \phi=(n,e,d,r), \qquad e: y=g(\mathbf{x}),
\end{equation}
where $n$ is its name, $e$ its equation, $d$ its physical description, and $r$ its applicability conditions.
The inputs $\mathbf{x}$ determine the target quantity $y$.
Extraction favors general relations that support multiple numerical instances.
An LLM plausibility filter checks equation well-formedness and consistency among the fields; deduplication then produces the formula library $\mathcal{F}$.
The library makes each relation and its physical scope explicit, providing conditions for downstream generation.
Source questions and answers serve only as extraction material and are not included as training examples; the training problems are generated through formula instantiation.

\subsection{Top-Down Formula Instantiation}
\label{sec:instantiation}

\paragraph{Program synthesis and validation.}
For each $\phi$, the \emph{Program Synthesiser} constructs an executable program $P_\phi$ mapping the independent inputs to $y$, together with a test set $T_\phi$.
Physical constants are incorporated into the program, and its input/output meanings and units are documented.
Test inputs lie within the formula's applicable domain.
Expected outputs are calculated from explicit formula-derived expressions without calling the program under test.
\emph{Program Validation} retains a program only if all tests pass:
\begin{equation}
    \operatorname{Keep}(P_\phi)
    =
    \mathbb{1}\!\left[
        \text{all tests in } T_\phi \text{ pass}
    \right].
\end{equation}
This checks computational consistency on the tested inputs.
The tests assess the implementation on designated formula-derived cases; they precede the execution of the parameter assignments used to construct training problems.
The retained program supplies the interface for parameter generation and the implementation for numerical execution; subsequent uses of $P_\phi$ refer to this retained program.

\paragraph{Parameter generation and instance execution.}
Conditioned on the formula's name, equation, applicability conditions, and program interface, the \emph{Parameter Generator} proposes input assignments.
Instructions require physically admissible values, relevant variable constraints, and valid numerical operations.
PHYSICS-derived assignments are used directly, while SciBench-derived assignments undergo rule-based perturbation to increase numerical diversity.
We denote the final assignment by $\theta$ and use it consistently for execution and question construction.
\emph{Instance Execution} computes
\begin{equation}
    y^* = P_\phi(\theta).
\end{equation}
Instances that raise execution errors are discarded.
Both source-specific instantiation routes therefore produce the same interface for downstream modules: an assignment paired with a successfully executed program and its numerical output.
The reference $y^*$ is fixed before question generation and teacher inference.
Repeated instantiation yields different configurations under the same physical relation, with generation controlled through formula selection and parameter constraints.
The full perturbation rules are given in Appendix~\ref{app:data_construction}.

\paragraph{Question verbalization.}
The \emph{Question Verbalizer} uses the formula information, retained program, $\theta$, and $y^*$ to generate an English problem $q$.
Both parameter generation and verbalization withhold the extracted description $d$ to limit source-specific context.
The problem must preserve the supplied values, describe variables through their physical meanings, and specify a complete scenario consistent with the applicability conditions.
It must provide enough information to recover the inputs and solve for the target while omitting the target equation.
The reference answer constrains problem construction, while the final statement presents the physical scenario and the quantities needed for solving it.
The verbalizer also supplies an auxiliary solution constrained to yield $y^*$ and identifies the answer unit.
This solution supports problem construction; training traces are obtained separately from the teacher.
SciBench-derived problems undergo additional plausibility filtering, while PHYSICS-derived problems undergo LaTeX normalization before teacher inference, as detailed in Appendix~\ref{app:data_construction}.

\subsection{Teacher Distillation}
\label{sec:teacher_distillation}

\paragraph{Candidate generation.}
A reasoning teacher independently solves each generated problem, receiving the problem and answer-format instructions but not $y^*$.
We use \href{https://huggingface.co/Qwen/Qwen3-235B-A22B-Thinking-2507}{\defaultteacher} \citep{Qwen3} as the default distillation teacher for both PHYSICS- and SciBench-derived problems.
This teacher is used for TIRE-19k and all TIRE experiments unless an alternative teacher is explicitly specified in the teacher-sensitivity study (Section~\ref{sec:teacher_sensitivity}).
The teacher is queried once per problem ($K=1$), producing a candidate trace $\tau$ from which a final answer $\hat y$ is extracted.
A request is re-issued, up to three attempts, only when the call fails or returns no complete final answer; incorrectly answered problems are not re-sampled.
Source-specific prompts and sampling settings are provided in Appendices~\ref{app:prompt_generation} and~\ref{app:teacher_sampling}, respectively.

\paragraph{Answer verification and data selection.}
\emph{Answer Verification} compares each candidate answer with the execution-derived reference.
For scalar answers, acceptance is determined by
\begin{equation}
    \label{eq:answer_acceptance}
    A(\hat{y},y^*)
    =
    \mathbb{1}\!\left[
        |\hat{y}-y^*|
        \leq
        \epsilon \max\!\left(|\hat{y}|,|y^*|\right)
    \right],
\end{equation}
with $\epsilon=0.05$ for both sources.
For vector-valued answers, the criterion is applied elementwise.
Verification checks final numerical agreement; individual reasoning steps are not separately verified.
Problems without an accepted trace are excluded, so answer verification filters problems rather than selecting among traces.
After deduplicating by problem statement, we randomly select one accepted trace per retained problem.
Answer extraction and final cleanup details appear in Appendix~\ref{app:data_construction}.

\paragraph{Dataset and training use.}
After the training split and final formatting cleanup, TIRE-19k contains 19,144 training examples associated with 3,051 formula entries.
Each example pairs a generated problem with an execution-derived reference and a selected teacher trace; Table~\ref{tab:tide_composition} reports the source composition.
The selected trace supplies a worked response for the generated question, and the reference remains the numerical target established before teacher sampling.
For SFT, the target response contains the teacher's reasoning followed by its answer.
For GRPO \citep{DeepSeekMath}, $y^*$ provides the reward reference, and agreement under the same tolerance gives a binary correctness reward.
Student training settings are described in Appendix~\ref{app:training_configurations}.

\section{Dataset Analysis}
\label{sec:dataset_analysis}

\begin{table}[t]
    \centering
    \fontsize{10pt}{12pt}\selectfont
    \caption{Composition of the TIRE-19k training set. Formula counts are deduplicated within each source. Two records occur in both sources, so the total is their globally deduplicated union.}
    \label{tab:tide_composition}
    \begin{tabular}{llrr}
        \toprule
        Formula source & Instantiation & Formulas & Examples \\
        \midrule
        PHYSICS  & Direct    & 2,561 & 15,133 \\
        SciBench & Perturbed &   492 &  4,011 \\
        Total    &           & 3,051 & 19,144 \\
        \bottomrule
    \end{tabular}
\end{table}

\begin{figure}[h!]
    \centering
    \includegraphics[width=\textwidth]{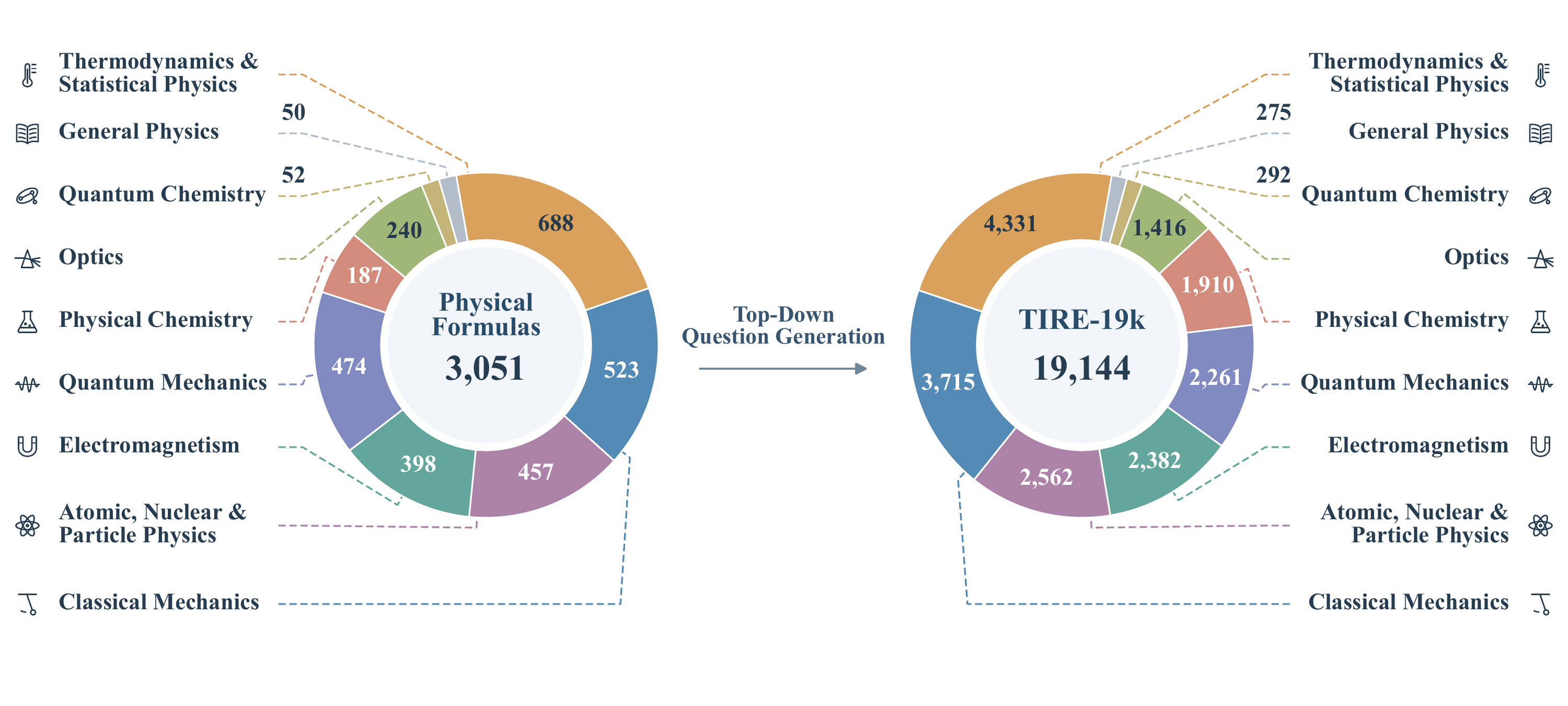}
    \caption{Subject distributions of formula records associated with TIRE-19k and training examples. Formula records can belong to multiple subjects, giving 3,069 subject memberships for 3,051 globally unique records; formula percentages use subject memberships as the denominator.}
    \label{fig:tide_subject_distribution}
\end{figure}

\begin{figure}[!htbp]
    \centering
    \includegraphics[width=\textwidth]{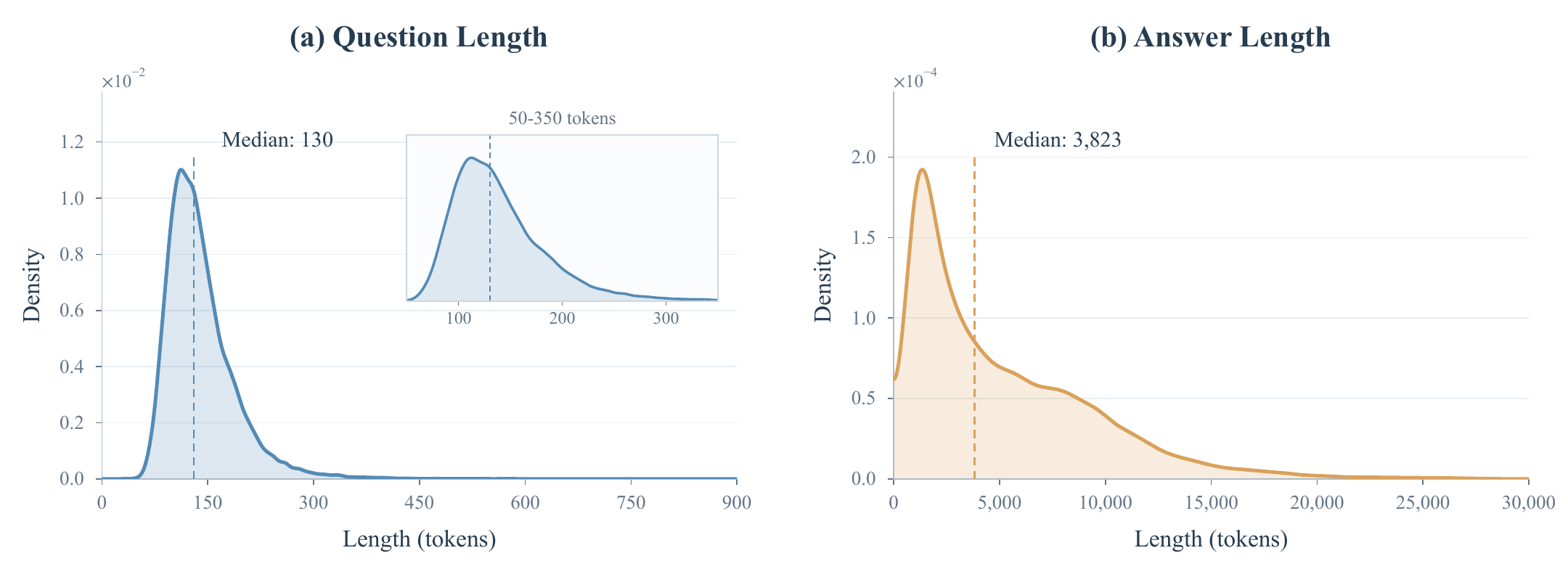}
    \caption{Question and answer length distributions in TIRE-19k. Dashed lines mark medians.}
    \label{fig:tide_length_distribution}

    \par\vspace{8pt}
    \includegraphics[width=\textwidth]{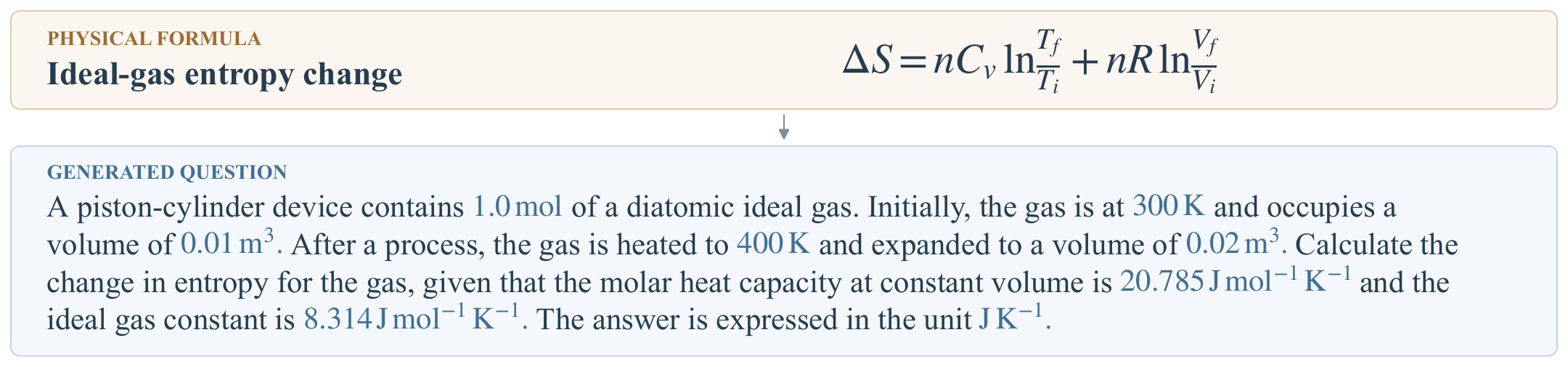}
    \caption{An example of formula-to-question generation in TIRE.}
    \label{fig:tide_case}
\end{figure}

\paragraph{Scale and source composition.}
TIRE-19k comprises 19,144 generated problems grounded in 3,051 formula records.
The formulas are drawn from PHYSICS and SciBench, with source composition summarized in Table~\ref{tab:tide_composition}.
The dataset averages 6.27 retained examples per formula, expanding reusable physical relations into multiple concrete problem instances.

\paragraph{Broad subject coverage.}
Figure~\ref{fig:tide_subject_distribution} shows coverage of nine subjects spanning classical and modern physics, physical chemistry, and quantum chemistry.
Seven subjects each contribute more than 1,400 examples, and the largest accounts for 22.62\% of the dataset.
Four subjects, including classical and quantum mechanics, each contain more than 450 formula records, providing a substantial knowledge base across domains.

\paragraph{Question and answer lengths.}
Figure~\ref{fig:tide_length_distribution} shows length distributions measured with the Qwen3-8B tokenizer.
Questions are compact, with a mean of 140 tokens and a median of 130; 99\% contain at most 309 tokens.
Teacher responses, including reasoning traces and final answers, average 5,136 tokens with a median of 3,823, and 21.94\% exceed 8,192 tokens.
This pairing of concise questions with extended worked responses provides substantial reasoning supervision across a wide range of response lengths.

\paragraph{Illustrative example.}
Figure~\ref{fig:tide_case} illustrates the instantiation of the ideal-gas entropy-change formula as a piston-cylinder problem.
The generated question specifies the gas amount, initial and final states, heat capacity, and gas constant, expressing an abstract relation as a concrete physical scenario with a clear target quantity and answer unit.


\section{Experiments}
\label{sec:experiments}

We train 3 backbone LLMs on TIRE-19k and established physics datasets using the same number of training examples.
We evaluate the resulting models on 4 representative physics benchmarks.

\subsection{Experimental Setup}
\label{sec:experimental_setup}

The comparison spans three backbones: Qwen3-8B \citep{Qwen3}, DeepSeek-R1-Distill-Qwen-7B \citep{DeepSeek-R1}, and Qwen3.5-9B.\footnote{\url{https://huggingface.co/Qwen/Qwen3.5-9B}}
For each backbone, we compare TIRE-19k with five baseline datasets: MegaScience \citep{MEGAScience}, PhysLogic \citep{PhysLogic}, Spark-234K (Spark) \citep{SPARK}, OpenSciReasoning-Physics-20K (OpenSci),\footnote{\url{https://huggingface.co/datasets/TerryJCZhang/OpenSciReasoning-Physics-20K}} and SCP-116k \citep{SCP-116K}.
Each comparison uses 19,144 training examples per dataset.
The main SFT settings are detailed in Appendix~\ref{app:training_configurations}.

Evaluation covers in-domain and out-of-domain performance.
SciBench Physics \citep{SciBench} is in-domain because it contributes to TIRE's formula library; PhysReason \citep{PhysReason}, OlympiadBench Physics \citep{OlympiadBench}, and TheoremQA-Physics \citep{TheoremQA} assess generalization beyond TIRE's sources.
We report benchmark accuracies and their unweighted average as percentages; differences are computed from the displayed scores.

Component ablations train Qwen3.5-9B on four dataset configurations: full TIRE, TIRE without unit tests, TIRE without parameter perturbation, and Self-QA (\emph{w/o all}), which generates question--answer pairs directly from the same formula records.
Each ablation dataset contains 19,144 examples, and all configurations use the default teacher, \defaultteacher; detailed ablation settings are given in Appendix~\ref{app:ablation_data}.

We assess teacher sensitivity with Qwen3.5-9B as the student, using the default \defaultteacher\ and three alternative teachers: Qwen3.7-plus, Qwen3.8-flash, and DeepSeek-v4.1-flash.

To assess further improvement after SFT, we continue training Qwen3-8B with GRPO on MegaScience, PhysLogic, and TIRE-19k.
Each run starts from the corresponding SFT checkpoint and uses the same dataset, with numerical correctness as the reward.

Evaluation details appear in Appendix~\ref{app:evaluation_protocol}.

\subsection{SFT Experiments}
\label{sec:comparison}

\begin{table}[t]
    \centering
    \fontsize{10pt}{12pt}\selectfont
    \captionsetup{position=top,skip=4pt}
    \caption{
        Comparison of training datasets on Qwen3.5-9B.
        All scores are percentages.
    }
    \label{tab:comparison_qwen3_5_9b}
    \begin{tabular}{lccccc}
        \toprule
        Training data & SciBench & PhysReason & OlympiadBench
        & TheoremQA & Average \\
        \midrule
        MegaScience
        & 56.71 & 66.17 & 26.69 & 32.46 & 45.51 \\
        PhysLogic
        & 66.97 & 68.58 & 26.69 & 64.04 & 56.57 \\
        Spark
        & 74.56 & 74.31 & 43.64 & 65.79 & 64.58 \\
        OpenSci
        & 78.84 & 82.26 & \underline{47.88} & 64.04 & 68.26 \\
        SCP-116k
        & \underline{79.22} & \underline{82.62} & 43.64
        & \underline{69.30} & \underline{68.70} \\
        \rowcolor{tideblue}
        TIRE-19k
        & \textbf{83.76} & \textbf{82.99} & \textbf{51.27}
        & \textbf{71.05} & \textbf{72.27} \\
        \bottomrule
    \end{tabular}

    \par\vspace{14pt}
    \caption{
        Comparison of training datasets on Qwen3-8B.
        All scores are percentages.
    }
    \label{tab:comparison_qwen3_8b}
    \begin{tabular}{lccccc}
        \toprule
        Training data & SciBench & PhysReason & OlympiadBench
        & TheoremQA & Average \\
        \midrule
        MegaScience
        & 44.41 & 62.29 & 19.49 & 44.74 & 42.73 \\
        PhysLogic
        & 61.01 & 62.85 & 23.31 & 52.63 & 49.95 \\
        Spark
        & 56.03 & 68.02 & 22.88 & 50.88 & 49.45 \\
        OpenSci
        & \underline{72.49} & \underline{76.71} & 37.71
        & 61.40 & 62.08 \\
        SCP-116k
        & 71.64 & 75.79 & \underline{39.83}
        & \underline{67.54} & \underline{63.70} \\
        \rowcolor{tideblue}
        TIRE-19k
        & \textbf{78.08} & \textbf{78.19} & \textbf{40.68}
        & \textbf{69.30} & \textbf{66.56} \\
        \bottomrule
    \end{tabular}

    \par\vspace{14pt}
    \caption{
        Comparison of training datasets on DeepSeek-R1-7B.
        All scores are percentages.
    }
    \label{tab:comparison_deepseek_r1_distill_qwen_7b}
    \begin{tabular}{lccccc}
        \toprule
        Training data & SciBench & PhysReason & OlympiadBench
        & TheoremQA & Average \\
        \midrule
        MegaScience
        & 61.05 & 57.49 & 15.68 & 48.25 & 45.62 \\
        PhysLogic
        & 60.18 & 52.50 & 15.25 & 39.47 & 41.85 \\
        Spark
        & 57.06 & 63.40 & 15.25 & 35.09 & 42.70 \\
        OpenSci
        & 54.07 & \underline{68.21} & 18.64
        & \underline{57.02} & 49.49 \\
        SCP-116k
        & \underline{65.67} & \textbf{71.72} & \textbf{27.54}
        & 52.63 & \underline{54.39} \\
        \rowcolor{tideblue}
        TIRE-19k
        & \textbf{68.09} & \textbf{71.72} & \underline{26.69}
        & \textbf{59.65} & \textbf{56.54} \\
        \bottomrule
    \end{tabular}

\end{table}

Tables~\ref{tab:comparison_qwen3_5_9b}, \ref{tab:comparison_qwen3_8b}, and~\ref{tab:comparison_deepseek_r1_distill_qwen_7b} show that TIRE-19k provides advantages over existing datasets in both in-domain performance and out-of-domain generalization across all three backbones.
On in-domain SciBench Physics, it exceeds the best baseline by 5.59, 2.42, and 4.54 percentage points on Qwen3-8B, DeepSeek-R1-Distill-Qwen-7B, and Qwen3.5-9B, respectively.

The gains also extend to the three out-of-domain benchmarks: PhysReason, OlympiadBench Physics, and TheoremQA-Physics.
Across these benchmarks, TIRE-19k's unweighted average accuracy exceeds that of the strongest baseline, SCP-116k, by 1.67, 2.06, and 3.25 percentage points on Qwen3-8B, DeepSeek-R1-Distill-Qwen-7B, and Qwen3.5-9B, respectively.

TIRE-19k has the highest four-benchmark average on every backbone.
Its average accuracy exceeds that of the strongest baseline by 2.86, 2.15, and 3.57 percentage points on Qwen3-8B, DeepSeek-R1-Distill-Qwen-7B, and Qwen3.5-9B, respectively.
\textbf{At the same training data scale, TIRE's formula-based, execution-verified data support stronger in-domain performance and out-of-domain generalization across backbones.}

A PHYSICS-only variant, trained under the same settings after removing all SciBench-derived examples and resampling the remaining data to the same size, still outperforms all five baselines on every benchmark for Qwen3.5-9B and Qwen3-8B (Appendix~\ref{app:source_data}, Table~\ref{tab:physics_only}).
\textbf{TIRE's advantage therefore persists when all four benchmarks are out-of-domain with respect to the training-data construction source.}

\subsection{Ablation Study}
\label{sec:ablation}

\begin{table}[t]
    \centering
    \fontsize{10pt}{12pt}\selectfont
    \setlength{\tabcolsep}{4pt}
    \caption{
        Ablation study of TIRE on Qwen3.5-9B, all scores are percentages.
    }
    \label{tab:ablation}
    \begin{tabular}{llllll}
        \toprule
        Configuration & SciB. & PhysR. & Olympiad.
        & TheoremQA. & Average \\
        \midrule
        \rowcolor{tideblue}
        TIRE
        & \textbf{81.29} & \textbf{84.10} & \textbf{49.58}
        & \textbf{71.05} & \textbf{71.51} \\
        \quad w/o unit tests
        & 79.69\textsubscript{\textcolor{red}{(-1.60)}}
        & 82.99\textsubscript{\textcolor{red}{(-1.11)}}
        & 47.46\textsubscript{\textcolor{red}{(-2.12)}}
        & \underline{70.18}\textsubscript{\textcolor{red}{(-0.87)}}
        & 70.08\textsubscript{\textcolor{red}{(-1.43)}} \\
        \quad w/o perturbation
        & \underline{79.72}\textsubscript{\textcolor{red}{(-1.57)}}
        & \underline{83.18}\textsubscript{\textcolor{red}{(-0.92)}}
        & \underline{48.31}\textsubscript{\textcolor{red}{(-1.27)}}
        & 69.30\textsubscript{\textcolor{red}{(-1.75)}}
        & \underline{70.13}\textsubscript{\textcolor{red}{(-1.38)}} \\
        \quad w/o all
        & 79.25\textsubscript{\textcolor{red}{(-2.04)}}
        & 81.33\textsubscript{\textcolor{red}{(-2.77)}}
        & 40.25\textsubscript{\textcolor{red}{(-9.33)}}
        & \underline{70.18}\textsubscript{\textcolor{red}{(-0.87)}}
        & 67.75\textsubscript{\textcolor{red}{(-3.76)}} \\
        \bottomrule
    \end{tabular}
\end{table}

Table~\ref{tab:ablation} shows that full TIRE achieves the highest accuracy on all four benchmarks, reaching an average of \textbf{71.51\%} with equal amounts of training data across configurations.
Removing unit tests lowers average accuracy by \textbf{1.43} percentage points, with the largest decrease on OlympiadBench (\textbf{2.12} points).
Removing parameter perturbation lowers average accuracy by \textbf{1.38} percentage points, with the largest decrease on TheoremQA-Physics (\textbf{1.75} points).
Both component ablations reduce accuracy on every benchmark, showing that their contributions extend to both in-domain performance and out-of-domain generalization.

Self-QA (\emph{w/o all}) uses the same formula records to generate problems and reference answers directly with a teacher, without program synthesis or execution.
TIRE instantiates formula-based programs and uses their outputs as independent reference answers.
Its gains of \textbf{3.76} percentage points in average accuracy and \textbf{9.33} points on OlympiadBench highlight the benefit of program-based instantiation and execution-derived reference answers over direct teacher-generated QA.
\textbf{These ablations support TIRE's instantiation and verification components.}

\subsection{Sensitivity to the Teacher LLM}
\label{sec:teacher_sensitivity}

Table~\ref{tab:teacher_sensitivity} shows that TIRE maintains strong performance across all four distillation teachers.
Average accuracy ranges from 69.84\% to 72.27\%, with all four teachers exceeding the strongest baseline average of 68.70\% in Table~\ref{tab:comparison_qwen3_5_9b}.
The default teacher achieves the highest average accuracy, exceeding the strongest alternative by 0.56 percentage points.
\textbf{TIRE remains effective with multiple distillation teachers, with the default teacher yielding the strongest overall results.}

\begin{table}[t]
    \centering
    \fontsize{10pt}{12pt}\selectfont
    \setlength{\tabcolsep}{3.5pt}
    \caption{
        Sensitivity to the distillation teacher with Qwen3.5-9B as the student.
    }
    \label{tab:teacher_sensitivity}
    \begin{tabular}{p{0.26\textwidth}ccccc}
        \toprule
        Teacher & SciBench & PhysReason & OlympiadBench
        & TheoremQA & Average \\
        \midrule
        Qwen3-235B-A22B
        & 83.76 & 82.99 & \textbf{51.27} & \textbf{71.05} & \textbf{72.27} \\
        Qwen3.7-plus
        & 81.76 & 84.66 & 47.46 & \textbf{71.05} & 71.23 \\
        Qwen3.8-flash
        & \textbf{84.52} & 85.58 & 48.31 & 68.42 & 71.71 \\
        DeepSeek-v4.1-flash
        & 82.74 & \textbf{85.95} & 46.61 & 64.04 & 69.84 \\
        \bottomrule
    \end{tabular}
\end{table}

\subsection{Further Training with GRPO}
\label{sec:grpo}

\begin{table}[t]
    \centering
    \fontsize{10pt}{12pt}\selectfont
    \setlength{\tabcolsep}{3pt}
    \caption{
        Qwen3-8B accuracy (\%) after SFT and after further GRPO, with Average subscripts indicating absolute accuracy changes in percentage points from the corresponding SFT checkpoint.
    }
    \label{tab:sft_grpo}
    \begin{tabular}{@{}llccccl@{}}
        \toprule
        Training data & Setting & SciBench & PhysReason
        & Olympiad. & TheoremQA & Average \\
        \midrule
        \multirow{2}{*}{MegaScience}
        & SFT
        & 44.41 & 62.29 & 19.49 & 44.74 & 42.73 \\
        & SFT$\rightarrow$GRPO
        & 46.09 & 65.25 & 16.95 & 42.11 & 42.60\textsubscript{\textcolor{red}{(-0.13)}} \\
        \addlinespace
        \multirow{2}{*}{PhysLogic}
        & SFT
        & 61.01 & 62.85 & 23.31 & 52.63 & 49.95 \\
        & SFT$\rightarrow$GRPO
        & 62.38 & 62.85 & 19.92 & 50.88 & 49.01\textsubscript{\textcolor{red}{(-0.94)}} \\
        \addlinespace
        \rowcolor{tideblue}
        & SFT
        & \underline{78.08} & \textbf{78.19} & \underline{40.68} & \underline{69.30} & \underline{66.56} \\
        \rowcolor{tideblue}
        \multirow{-2}{*}{TIRE-19k}
        & SFT$\rightarrow$GRPO
        & \textbf{78.64} & \underline{77.63} & \textbf{43.22} & \textbf{72.81} & \textbf{68.08}\textsubscript{\textcolor{green!50!black}{(+1.52)}} \\
        \bottomrule
    \end{tabular}
\end{table}

On Qwen3-8B, further GRPO raises TIRE-19k's average accuracy from 66.56\% after SFT to 68.08\%, yielding the largest post-SFT average gain of \textbf{1.52} percentage points, versus $-0.13$ for MegaScience and $-0.94$ for PhysLogic (Table~\ref{tab:sft_grpo}).
The improvement spans in-domain and out-of-domain evaluation: TIRE-19k gains \textbf{0.56} percentage points on SciBench Physics, as well as \textbf{2.54} on OlympiadBench Physics and \textbf{3.51} on TheoremQA-Physics, where both baselines decline.
\textbf{TIRE-19k is the most effective of the compared datasets for further GRPO training, supporting additional gains after SFT on both in-domain and out-of-domain benchmarks.}

\section{Conclusion}

TIRE constructs physics reasoning data through formula-guided problem instantiation and execution-based verification of teacher answers.
With equal amounts of SFT data, TIRE-19k outperforms existing datasets in both in-domain performance and out-of-domain generalization, achieving the highest four-benchmark average accuracy across all three backbones.
Further gains after SFT through GRPO and effectiveness across teachers support its use for scalable physics training-data construction.

\bibliography{references}
\bibliographystyle{references}

\appendix

\section{Data Construction Details}
\label{app:data_construction}

\subsection{Source Data and Formula Library}
\label{app:source_data}

Formulas are extracted from PHYSICS and SciBench; each extraction request contains a source problem together with its official solution.

\paragraph{PHYSICS.}
We use the official release of PHYSICS~\citep{PHYSICS},\footnote{\url{https://github.com/yale-nlp/Physics/tree/main/PHYSICS}} which contains 1,297 problems drawn from publicly available physics PhD qualifying examinations, organized in six subject files (\texttt{atomic}, \texttt{electro}, \texttt{mechanics}, \texttt{optics}, \texttt{quantum}, and \texttt{statistics}).
Formula extraction covers the 999 text-only problems, and PHYSICS is not used for evaluation.

\paragraph{SciBench.}
SciBench~\citep{SciBench}\footnote{\url{https://github.com/mandyyyyii/scibench/tree/main/dataset/original}} collects college-level problems from ten textbooks in physics, chemistry, and mathematics.
Its three physics textbooks, \texttt{fund}, \texttt{class}, and \texttt{thermo} (193 problems), form the SciBench Physics evaluation set and are also used for formula extraction, together with the textbooks of other disciplines whose problems apply physical laws; the mathematics textbooks are excluded.
Both benchmarks serve here as sources of physical knowledge rather than of training problems: following Physics Reasoner~\citep{PhysicsReasoner}, which collects formulas from the same three physics textbooks, we use SciBench for the laws of college physics curricula, and PHYSICS for those required at the level of PhD qualifying examinations.
Because the evaluation textbooks contribute formulas, SciBench Physics is treated as in-domain for the full TIRE-19k dataset; the PHYSICS-only comparison below evaluates performance when none of the evaluation benchmarks is used as a construction source.

\paragraph{Formula records.}
Records are identified by their name and formula string, so algebraic variants of the same law, such as $F=ma$ and $a=F/m$, are distinct records: they have different target variables and therefore yield different programs and problems.
After the plausibility filter, near-duplicate records within each source are removed: candidates are retrieved with MinHash locality-sensitive hashing over names and normalized formula strings, and a record is dropped when its name or formula nearly matches a retained record (Levenshtein ratio at least 0.82 or edit distance at most 5).
Each TIRE-19k example is anchored to the formula record whose program computes its reference answer; solving the problem can still require further relations, such as unit conversions, physical constants, or intermediate quantities.
Over the final data, PHYSICS-derived examples are linked to 2,561 records and SciBench-derived examples to 492, for 3,051 in total since two records occur in both sources (Table~\ref{tab:tide_composition}).

\paragraph{Evaluation with PHYSICS-only training data.}
To test whether TIRE-19k's advantages persist without using evaluation benchmarks as construction sources, we remove all 4,011 SciBench-derived examples and supplement the remaining 15,133 PHYSICS-derived examples by resampling them to restore the original size of 19,144.
We train Qwen3.5-9B and Qwen3-8B on this PHYSICS-only variant with the same training and evaluation settings as for the full dataset.
None of the four evaluation benchmarks supplies problems or solutions to the construction of the retained PHYSICS-derived examples; all four benchmarks are therefore out-of-domain with respect to this variant's construction source.

\begin{table}[htbp]
    \centering
    \fontsize{10pt}{12pt}\selectfont
    \setlength{\tabcolsep}{4pt}
    \caption{Evaluation after excluding SciBench-derived training data. Both variants contain 19,144 examples before the validation split. All scores are percentages; Average is the unweighted mean of the four benchmark accuracies.}
    \label{tab:physics_only}
    \begin{tabular}{lccccc}
        \toprule
        Construction source & SciBench & PhysReason & Olympiad. & TheoremQA & Average \\
        \midrule
        \multicolumn{6}{l}{\textbf{Qwen3.5-9B}} \\
        PHYSICS + SciBench & 83.76 & 82.99 & 51.27 & 71.05 & 72.27 \\
        PHYSICS only & 82.01 & 83.73 & 48.73 & 72.81 & 71.82 \\
        \addlinespace
        \multicolumn{6}{l}{\textbf{Qwen3-8B}} \\
        PHYSICS + SciBench & 78.08 & 78.19 & 40.68 & 69.30 & 66.56 \\
        PHYSICS only & 74.98 & 80.59 & 40.68 & 71.05 & 66.82 \\
        \bottomrule
    \end{tabular}
\end{table}

Table~\ref{tab:physics_only} shows that PHYSICS-only training achieves average accuracies of 71.82\% and 66.82\% on Qwen3.5-9B and Qwen3-8B, respectively.
The PHYSICS-only variant outperforms all five baseline datasets on every benchmark for both backbones (Tables~\ref{tab:comparison_qwen3_5_9b} and~\ref{tab:comparison_qwen3_8b}), with its four-benchmark average exceeding the strongest baseline by 3.12 points in each case.
Its average accuracy remains within 0.45 percentage points of the full configuration on both backbones.
These results show that TIRE's performance advantage over the compared datasets persists when none of the evaluation benchmarks is used as a construction source.
We include SciBench to broaden TIRE-19k's curricular coverage and data diversity.
Its undergraduate-level physical knowledge complements the graduate-level knowledge drawn from PHYSICS qualifying examinations.

\subsection{Construction Configuration}
\label{app:construction_configurations}

Table~\ref{tab:construction_configurations} lists the construction stages and settings for DeepSeek-R1-250528,\footnote{\url{https://huggingface.co/deepseek-ai/DeepSeek-R1-0528}} DeepSeek-V3.1,\footnote{\url{https://huggingface.co/deepseek-ai/DeepSeek-V3.1}} Doubao-1.6-thinking,\footnote{\url{https://seed.bytedance.com/en/blog/introduction-to-techniques-used-in-seed1-6}} and the distillation teacher described below.
Unspecified sampling parameters follow the defaults of the serving endpoint.

\begin{table}[htbp]
    \centering
    \fontsize{10pt}{12pt}\selectfont
    \caption{Construction configuration of TIRE-19k.}
    \label{tab:construction_configurations}
    \begin{tabular}{p{0.27\textwidth}>{\raggedright\arraybackslash}p{0.66\textwidth}}
        \toprule
        Stage & Model and main settings \\
        \midrule
        Formula extraction & DeepSeek-R1-250528; at most 16,000 output tokens \\
        \addlinespace[8pt]
        Formula filtering & DeepSeek-R1-250528; True/False judgment; temperature 0.6 \\
        \addlinespace[8pt]
        Program and unit-test generation & DeepSeek-V3.1, thinking mode; temperature 0.6 \\
        \addlinespace[8pt]
        Parameter generation & DeepSeek-V3.1, thinking mode; temperature 0.6 \\
        \addlinespace[8pt]
        Parameter perturbation & Rule-based, $\pm5\%$ \\
        \addlinespace[8pt]
        Question verbalization & DeepSeek-V3.1, thinking mode; temperature 0.6; at most 20,000 output tokens \\
        \addlinespace[8pt]
        Problem filtering & Doubao-1.6-thinking; YES/NO judgment; temperature 0.1; SciBench-derived problems only \\
        \addlinespace[8pt]
        LaTeX normalization & Doubao-1.6-thinking; temperature 0.1; PHYSICS-derived problems only \\
        \addlinespace[8pt]
        Teacher inference & Qwen3-235B-A22B, thinking mode; temperature 0.6; one trace per problem \\
        \bottomrule
    \end{tabular}
\end{table}

\paragraph{Parameter perturbation.}
For SciBench-derived instances, every nonzero scalar parameter $v$ is replaced by $v(1+u)$ with $u\sim\mathcal{U}(-0.05,0.05)$ drawn independently, zero-valued scalars by a draw from $\mathcal{U}(0,0.0005)$, and list elements are perturbed in the same way except that zeros are kept; values are rounded to three decimal places unless rounding would make them zero.
Execution, verbalization, and verification use the perturbed assignment.

\paragraph{Problem filtering and LaTeX normalization.}
The problem filter reads only the generated problem statement and removes problems whose values are not physically meaningful (e.g., non-integer counts), whose units are inconsistent, or whose scenario violates basic scientific principles; it retains 4,553 of 5,911 SciBench-derived problems.
LaTeX normalization rewrites units, constants, and symbols in LaTeX without changing the problem, and the rewrite is kept only if its ROUGE-L F1\footnote{\url{https://aclanthology.org/W04-1013/}} with the original exceeds 0.5.

\subsection{Teacher Sampling and Data Selection}
\label{app:teacher_sampling}

\paragraph{Sampling.}
Each problem is sent once to \defaultteacher\footnote{\url{https://huggingface.co/Qwen/Qwen3-235B-A22B-Thinking-2507}} ($K=1$), with temperature 0.6 and at most 28,000 output tokens.
A request is re-issued only when the call fails or the response contains no complete final answer; problems answered incorrectly are not re-sampled, so answer verification acts as a filter on problems.

\paragraph{Answer verification.}
The answer is the content of the last \texttt{\textbackslash boxed\{\}} in the final response after the reasoning block, parsed as a number (with Math-Verify\footnote{\url{https://github.com/huggingface/Math-Verify}} as a fallback).
It is accepted if it satisfies Equation~\ref{eq:answer_acceptance} with $\epsilon=0.05$; list-valued answers are compared element by element.
The teacher prompt states the unit of the answer and asks for the boxed value without units, and no unit conversion is performed, so a value reported in another unit or scale is rejected.
Responses without a parseable boxed answer are rejected.

\paragraph{Final format.}
Each example is a chat conversation whose assistant turn contains the teacher's reasoning in \texttt{<think>} tags followed by its final response; only untruncated traces that pass answer verification are kept.

\section{Experimental Protocols}
\label{app:experimental_protocols}

\subsection{Training Configurations}
\label{app:training_configurations}

Table~\ref{tab:training_configurations} summarizes the default training settings, with ablation-specific settings in Appendix~\ref{app:ablation_data}; benchmark and decoding settings are given in Appendix~\ref{app:evaluation_protocol}.
We use ms-swift\footnote{\url{https://github.com/modelscope/ms-swift}} for SFT and GRPO, DeepSpeed ZeRO-2\footnote{\url{https://www.deepspeed.ai/tutorials/zero/}} for SFT, and vLLM\footnote{\url{https://github.com/vllm-project/vllm}} for GRPO rollouts.
The alternative teachers in Table~\ref{tab:teacher_sensitivity} are Qwen3.7-plus and Qwen3.8-flash,\footnote{\url{https://www.alibabacloud.com/help/en/model-studio/models}} and DeepSeek-v4.1-flash.\footnote{\url{https://huggingface.co/deepseek-ai/DeepSeek-V4.1-Flash}}

\begin{table}[htbp]
    \centering
    \fontsize{10pt}{12pt}\selectfont
    \caption{Training configurations.}
    \label{tab:training_configurations}
    \begin{tabular}{p{0.29\textwidth}p{0.64\textwidth}}
        \toprule
        Configuration & Setting \\
        \midrule
        SFT optimization & Full-parameter ms-swift, 5 epochs; AdamW ($\beta_1=0.9$, $\beta_2=0.95$); learning rate $10^{-6}$, cosine decay, 10\% warmup; weight decay 0.1; gradient clipping 1.0; global batch size 2; bf16, ZeRO-2, no packing; the epoch-5 checkpoint is evaluated \\
        \addlinespace[8pt]
        SFT data use & Main dataset comparisons: 2\% of each 19,144-example set is held out for validation (seed 42), leaving 18,762 training examples; the ablation datasets also contain 19,144 examples and use the shorter configuration in Appendix~\ref{app:ablation_data} \\
        \addlinespace[8pt]
        Maximum length and thinking mode & 40,960 with thinking (TIRE-19k, SCP-116k, and the Qwen3.7-plus teacher in Table~\ref{tab:teacher_sensitivity}); 20,480 with thinking (OpenSci; Qwen3.8-flash and DeepSeek-v4.1-flash teachers in Table~\ref{tab:teacher_sensitivity}); 16,384 without thinking (Spark); 8,192 without thinking (MegaScience) or with thinking (PhysLogic); rationale given below the table \\
        \addlinespace[8pt]
        GRPO (Table~\ref{tab:sft_grpo}) & Qwen3-8B; each run starts from the epoch-5 SFT checkpoint of its dataset and trains for 3 epochs on the prompts whose four-sample solve rate under that checkpoint lies in $(0,1)$; ms-swift GRPO with vLLM rollouts; binary reward (Equation~\ref{eq:answer_acceptance}); 8 rollouts per prompt, 128 completions per update; learning rate $10^{-6}$ (cosine, 10\% warmup); 16,384 tokens, at most 14,336 generated; temperature 0.6, top-p 0.95, top-k 80; KL $\beta=0.04$, clipping $\epsilon=0.2$ \\
        \addlinespace[8pt]
        Training hardware & SFT: 4 NVIDIA H200 GPUs (141 GB), some baselines on 2 GPUs at the same global batch size; ablations (Table~\ref{tab:ablation}): 8 H200 GPUs; GRPO: 8 H200 GPUs \\
        \bottomrule
    \end{tabular}
\end{table}

\paragraph{Dataset-dependent settings.}
For the main dataset comparisons, SFT runs within each backbone share the optimization settings and the number of examples; only the maximum length and the thinking mode depend on the dataset.
The maximum length is set from the longest example in the sampled data of each dataset.
When only a few examples form a long tail, they are replaced by shorter examples from the same dataset so that a shorter limit suffices (e.g., PhysLogic at 8,192); when long examples are common, as in TIRE-19k and SCP-116k, they are kept under a larger limit (40,960).
Datasets without reasoning traces (MegaScience and Spark) are fine-tuned with thinking disabled and evaluated in the same mode; all other datasets, as well as the backbones, use thinking mode.

\subsection{Ablation Datasets}
\label{app:ablation_data}

All four configurations in Table~\ref{tab:ablation} use datasets of 19,144 examples built with the construction configuration in Table~\ref{tab:construction_configurations} and the Qwen3.5-9B backbone, with \defaultteacher\ as the teacher.
Full TIRE serves as the common control for the two component ablations, each of which disables the corresponding step in the main construction pipeline while retaining the other construction stages.
To accommodate the shorter examples in the ablation datasets, all four configurations are trained with a maximum length of 24,576 tokens and a global batch size of 4; the other settings follow Table~\ref{tab:training_configurations}.

\paragraph{Unit tests.}
The \emph{w/o unit tests} variant omits unit-test-based program validation: generated programs proceed to parameter generation and execution without requiring their unit tests to pass.
Program execution still supplies the reference answers, and instances that raise execution errors or yield no usable reference answer are discarded.
Question verbalization, source-specific processing, teacher distillation, and final-answer verification otherwise follow the full pipeline.
The variant includes 2,256 verified examples from programs that fail their unit tests and is supplemented with examples sampled from TIRE-19k to reach 19,144 examples while preserving the source composition.

\paragraph{Parameter perturbation.}
The \emph{w/o perturbation} variant replaces the perturbed SciBench-derived examples with unperturbed ones: the generated assignments are used directly for program execution and question verbalization, without the rule-based perturbation of Appendix~\ref{app:construction_configurations}.
PHYSICS-derived assignments remain unperturbed in both full TIRE and this variant.

\paragraph{Self-QA.}
The \emph{w/o all} variant skips program synthesis, parameter generation, and execution: the teacher writes numerical problems directly from the same formula records and solves them with the formula in view to obtain the reference answers.
The teacher then solves each problem again from its statement alone, and a trace is retained only if its final answer agrees with the reference.

\subsection{Evaluation Protocols}
\label{app:evaluation_protocol}

Tables~\ref{tab:evaluation_benchmarks} and~\ref{tab:evaluation_settings} specify the benchmark subsets and evaluation configurations.
We use the official PhysReason data release,\footnote{\url{https://huggingface.co/datasets/zhibei1204/PhysReason}} OlympiadBench MathJudger,\footnote{\url{https://github.com/OpenBMB/OlympiadBench/blob/main/inference/code/math_judger.py}} and TheoremQA numerical scoring routines.\footnote{\url{https://github.com/TIGER-AI-Lab/TheoremQA/blob/main/number_utils.py}}
PhysReason responses are graded by DeepSeek-V4-Flash (0731 snapshot).\footnote{\url{https://huggingface.co/deepseek-ai/DeepSeek-V4-Flash-0731}}
For Qwen3-8B, context extension uses YaRN\footnote{\url{https://qwen.readthedocs.io/en/latest/deployment/vllm.html\#context-length}} with the scaling factor specified in Table~\ref{tab:evaluation_settings}.

\begin{table}[htbp]
    \centering
    \fontsize{10pt}{12pt}\selectfont
    \setlength{\tabcolsep}{4pt}
    \caption{Benchmark subsets, question counts, and scoring.}
    \label{tab:evaluation_benchmarks}
    \begin{tabular}{p{0.22\textwidth}>{\raggedright\arraybackslash}p{0.29\textwidth}p{0.11\textwidth}>{\raggedright\arraybackslash}p{0.27\textwidth}}
        \toprule
        Benchmark & Subset & Questions & Scoring \\
        \midrule
        SciBench & Physics: textbooks \texttt{fund}, \texttt{class}, and \texttt{thermo} of the official release & 193 & Numeric match with 5\% relative tolerance; macro-average over the three textbooks \\
        \addlinespace[8pt]
        PhysReason & Text-only questions with a single boxed reference answer & 541 & LLM judge (DeepSeek-V4-Flash, 0731 snapshot, thinking disabled) \\
        \addlinespace[8pt]
        OlympiadBench & Physics: \texttt{OE\_TO\_\allowbreak physics\_en\_COMP} (open-ended, text-only, English) & 236 & Official MathJudger with the per-question tolerance of the dataset \\
        \addlinespace[8pt]
        TheoremQA & Physics questions without images & 114 & Numeric match: 4\% relative tolerance for reals, exact for integers, lists compared after sorting \\
        \bottomrule
    \end{tabular}
\end{table}

\begin{table}[htbp]
    \centering
    \fontsize{10pt}{12pt}\selectfont
    \setlength{\tabcolsep}{3pt}
    \caption{Prompts, decoding, and answer processing.}
    \label{tab:evaluation_settings}
    \begin{tabular}{p{0.24\textwidth}>{\raggedright\arraybackslash}p{0.16\textwidth}>{\raggedright\arraybackslash}p{0.16\textwidth}>{\raggedright\arraybackslash}p{0.18\textwidth}>{\raggedright\arraybackslash}p{0.15\textwidth}}
        \toprule
        Setting & SciBench & PhysReason & OlympiadBench & TheoremQA \\
        \midrule
        Prompt & User prompt with unit sentence; zero-shot & System and user prompt; zero-shot & Official English open-ended prompt; zero-shot & System and user prompt; zero-shot \\
        \addlinespace[8pt]
        Reasoning mode & \multicolumn{4}{p{\dimexpr0.65\textwidth+6\tabcolsep\relax}}{Thinking enabled for the backbones and for models fine-tuned on reasoning traces, including all TIRE models; disabled for models fine-tuned on MegaScience or Spark.} \\
        \addlinespace[8pt]
        Temperature / top-p & Model defaults: 0.6 / 0.95 (Qwen3-8B, R1-7B), 1.0 / 1.0 (Qwen3.5-9B); presence penalty 0 & 0.6 / 0.95; presence penalty 1.0; judge temperature 0 & 0.6 / 0.95; presence penalty 1.0 & 0.6 / 0.95; presence penalty 1.0 \\
        \addlinespace[8pt]
        Output-token limit & \multicolumn{4}{p{\dimexpr0.65\textwidth+6\tabcolsep\relax}}{Raised stepwise up to 60,000 tokens (server context 65,536; YaRN factor 2 for Qwen3-8B), regenerating only unfinished responses; responses still truncated are regenerated up to twice at 60,000 tokens. The judge uses at most 32,768 tokens.} \\
        \addlinespace[8pt]
        Responses per question & \multicolumn{4}{p{\dimexpr0.65\textwidth+6\tabcolsep\relax}}{One response per question; accuracy is the percentage of correct responses.} \\
        \addlinespace[8pt]
        Answer extraction & Last \texttt{\textbackslash boxed\{\}} after the reasoning block; scientific-notation normalization & Final response after the reasoning block, passed to the judge & Last \texttt{\textbackslash boxed\{\}} after the reasoning block & Last \texttt{\textbackslash boxed\{\}} after the reasoning block; numbers and lists parsed \\
        \addlinespace[8pt]
        Tolerance & 5\% relative & Decided by the judge against the reference & Per-question tolerance from the dataset; $10^{-8}$ otherwise & 4\% relative for reals; exact after rounding for integers \\
        \addlinespace[8pt]
        Failure handling & \multicolumn{4}{p{\dimexpr0.65\textwidth+6\tabcolsep\relax}}{Empty, unparseable, truncated, or failed responses count as incorrect and remain in the denominator. A score is final only if the run is complete and both the truncation rate and the extraction-failure rate are at most 3\%.} \\
        \bottomrule
    \end{tabular}
\end{table}
\section{Prompt Templates}
\label{app:prompt_templates}

\newtcolorbox{promptblockbreakable}[1][Prompt]{%
    enhanced, breakable, sharp corners,
    colback=white, colframe=black, boxrule=\fboxrule,
    left=\fboxsep, right=\fboxsep, top=\fboxsep, bottom=\fboxsep,
    before skip=\smallskipamount, after skip=\smallskipamount,
    before upper={\setlength{\parindent}{0pt}\raggedright\textbf{#1}\par\smallskip}
}

\subsection{Data Generation Prompts}
\label{app:prompt_generation}

This section gives one prompt for each construction stage; placeholders such as \texttt{\{formula\}} are filled at run time.
Prompts marked as translated were written in Chinese and are rendered in English with their structure, constraints, and output formats unchanged; long in-prompt examples are abbreviated where noted.

\begin{promptblock}[Formula Extraction Prompt (translated)]
\raggedright
You are a professional physicist. Based on the problem and its solution, list all the general physics formulas involved. Output each formula in the following format:\par
\texttt{\{"name": "theorem name", "formula": "LaTeX formula", "description": "theorem description", "applicable\_range": "applicable range"\}}\par
Wrap the complete JSON array in a \texttt{json} code block, and write all fields in Chinese.\par
If you cannot solve the problem, still return at least one relevant formula.\par
\medskip
Problem: \texttt{\{problem\_text\}}\par
\medskip
Solution: \texttt{\{problem\_solutions\}}
\end{promptblock}

\begin{promptblock}[Formula Plausibility Filter Prompt (translated)]
\raggedright
\texttt{\{formula\_record\}} \emph{(the formula record as JSON inside a code block)}\par
\medskip
Based on the content of the JSON fields above, strictly judge their plausibility:\par
- name: the formula name; it must be general and apt.\par
- formula: must be an equation in LaTeX format, scientifically sound and free of syntax errors.\par
- description: roughly describes the meaning of the formula.\par
- applicable\_range: roughly specifies the scope in which the formula applies.\par
Judge comprehensively whether all four items are plausible and correspond to one another.\par
If all of them are plausible and consistent, return only True; if any item is implausible or inconsistent, return only False.\par
Do not explain the reason; return only True or False.
\end{promptblock}

\begin{promptblock}[Program Generation Prompt (translated)]
\raggedright
\#\# Identity\par
You are an expert in physics and physical chemistry and are proficient in Python.\par
\#\# Task\par
- Given the information of a physics or physical chemistry formula, generate a Python function that represents the formula. The inputs of the function are all the variables on the right-hand side of the formula, and the output is the variable on the left-hand side. Constants in the formula must not be inputs; keep them implicit inside the function.\par
\#\#\# Formula information\par
\texttt{\{"name": "\{name\}", "formula": "\{formula\}", "description": "\{description\}", "applicable\_range": "\{applicable\_range\}"\}}\par
\#\# Requirements\par
- Output Python code, and do not output any explanation outside the code block.\par
- The code must include type annotations for the input parameters and the return value.\par
- The code must include a docstring describing the function and its parameters.\par
- Put the generated code in a JSON object whose key is python\_code and whose value is the generated code.\par
\#\# Code requirements\par
- If a Python package is needed, import it inside the function, e.g., \texttt{def gravitational\_acceleration(a: float, G: float, M: float, r: float) -> float: import numpy as np}\par
\#\# Return format\par
- Return a standard JSON object containing the following field: \texttt{\{"python\_code": "" // the corresponding Python code\}}
\end{promptblock}

\begin{promptblockbreakable}[Unit-Test Generation Prompt (translated; example abbreviated)]
\#\# Identity\par
You are a rigorous physics expert, proficient in Python unit testing (unittest).\par
\#\# Goal\par
Based on the given ``formula information'' and the source code of the corresponding Python function, write high-quality unit-test code that verifies whether the function correctly implements the formula: the return value of the function should be equivalent to the physical quantity on the left-hand side of the formula, and the parameter list of the function should cover all independent variables on the right-hand side (physical constants in the formula are treated as internal constants and are not passed as parameters).\par
\#\#\# Formula information\par
\texttt{\{"name": \{name\}, "formula": \{formula\}, "description": \{description\}, "applicable\_range": \{applicable\_range\}\}}\par
\#\#\# Source code of the Python function under test\par
\texttt{\{python\_code\}}\par
\#\# Generation requirements\par
1. Output format: return only one valid JSON object whose only key is \texttt{"test\_cases"} and whose value is a string of pure Python test code. No other text, explanation, tags, backticks, or extra keys may appear.\par
2. Do not import or reference the module containing the function under test (no \texttt{import xxx} or \texttt{from xxx import} of the function itself); import only necessary standard libraries (usually unittest and math; do not import them if unused).\par
3. Call the function by name directly, as if it were defined above in the same file.\par
4. Use physically reasonable values: correct orders of magnitude, and no non-physical values such as negative masses or negative absolute temperatures.\par
5. Principles for computing expected values (key point; hard-coding values by intuition is strictly forbidden):\par
\hspace*{1em}- Every expected value must be computed on the fly from the structure of the formula; never write guessed approximate numbers (no magic numbers).\par
\hspace*{1em}- If the formula contains exponentials, logarithms, powers, or roots, rebuild the theoretical value with expressions such as \texttt{math.exp}, \texttt{math.log}, or \texttt{**}; do not write truncated decimal approximations.\par
\hspace*{1em}- For half-life scenarios, write \texttt{expected = N0 * 0.5**n} or \texttt{N0 * math.exp(-lambda\_ * t)}; do not hard-code division constants such as \texttt{N0/1024} unless the expression itself naturally produces them (e.g., two half-lives may be written as \texttt{N0 * 0.5**2}; prefer this form).\par
\hspace*{1em}- ``Approximately unchanged'' tests (e.g., with a very small decay constant) must first compute expected from the formula and then assert actual against expected; never add a hand-estimated number from a source other than expected (e.g., 19999.999).\par
\hspace*{1em}- Do not add a second assertion that artificially fits the initial value merely to be ``close''; to show that a change is tiny, \texttt{self.assertLess(abs(expected - N0)/N0, 1e-5)} may be added.\par
6. Within one test, after asserting a return value against its formula-based expected value, do not assert the same return value again against a hand-written approximate constant.\par
7. Floating-point assertions: prefer \texttt{self.assertAlmostEqual(expected, actual, places=...)}, with the number of places chosen by magnitude (usually 6--10); for values spanning several orders of magnitude, use \texttt{rel = abs(expected - actual)/max(1.0, abs(expected))} and assert that rel is below a reasonable threshold, with justification.\par
8. If constants (such as G, c, or k\_B) are not built into the function but are needed for the expected value, define them in the test with the same values and compute the expected value consistently with the structure of the formula.\par
9. Do not use randomness, I/O, networking, or third-party libraries.\par
10. If the formula contains powers, roots, exponentials, or logarithms, give an explicit physically derived expression for the expected value in the test instead of a hard-coded magic number.\par
11. Class name: Test + function name in upper camel case. Test method names: test\_description.\par
\#\# Quality and self-check (performed internally, not written to the output)\par
- Check for hard-coded constants that do not correspond directly to the structure of the formula; replace any with symbolic expressions.\par
- Check for the pattern ``repeated assertion + manual estimate''; if present, remove the manual estimate.\par
- Check that every expected value is defined independently of the function under test (using the function under test to derive its own expected value is forbidden).\par
\#\# Strict output specification\par
- Output no extra text, such as ``Here is'', ``Below are'', ``Explanation'', \texttt{<artifact>}, code fences, or ``I'll create a set of test cases for the Hubble's Law function. Here's a JSON object containing the test cases''.\par
- Output nothing outside the JSON.\par
- The test\_cases string in the JSON may contain line breaks and indentation.\par
\#\#\# Example unit test (do not copy the example; rewrite it for the current function)\par
\emph{[One JSON example whose test\_cases value is a unittest class for a photon-sphere-radius function; omitted here for space.]}\par
\#\# Failure handling\par
If information is missing and tests cannot be constructed, still return valid JSON: put a minimal unittest structure in test\_cases and mark it with \texttt{self.fail('insufficient data')} in setUp, but first try to generate tests normally.
\end{promptblockbreakable}

\begin{promptblockbreakable}[Parameter Generation Prompt (translated)]
\#\# Role\par
You are an expert in physical chemistry and scientific computing with Python.\par
\#\# Task\par
We provide the structured information of a physics or physical chemistry formula and its Python implementation. Based on the function signature and the physical meaning of the formula, construct one set of test input parameters that ``can pass and are physically reasonable'', and return only JSON. The return value of the function corresponds to the physical quantity on the left-hand side of the formula; all arguments of the function correspond to the variables on the right-hand side (and variables implicitly required by the implementation). Constants used in the code need not appear in the test input.\par
\#\#\# Formula Information\par
\texttt{\{"name": "\{name\}", "formula": "\{formula\}", "applicable\_range": "\{applicable\_range\}"\}}\par
\#\#\# Python Function\par
\texttt{\{function\_code\}}\par
\#\# Generation requirements\par
1. Read only the arguments in the function signature and generate a reasonable physical value for each of them; do not omit any argument; avoid 0 or unreasonable extreme values unless the physics or the function logic explicitly allows them.\par
2. If an argument should be an array, vector, or matrix (evidence: the function body contains shape operations, slicing, broadcasting, or elementwise operations, or the variable is passed to \texttt{np.array}, \texttt{np.asarray}, \texttt{np.stack}, \texttt{np.concatenate}, or array arithmetic), represent it as a list, since it cannot otherwise be stored as JSON; element values must be physically reasonable, and the list should be short (2--4 elements).\par
3. If an argument denotes:\par
\hspace*{1em}- temperature: use a reasonable range of 250--1500 K (adjustable to the applicable domain of the formula);\par
\hspace*{1em}- pressure: use $10^{3}$--$10^{7}$ Pa (or follow the formula description);\par
\hspace*{1em}- length (m): avoid 0 and negative values;\par
\hspace*{1em}- time (s): use positive values;\par
\hspace*{1em}- concentration or fraction: use the interval (0, 1];\par
\hspace*{1em}- energy, enthalpy, entropy, and similar quantities: use signs and magnitudes consistent with common physical situations;\par
\hspace*{1em}- frequency or rate constant: use positive values;\par
\hspace*{1em}- names containing ref, std, or baseline: common standard-condition values may be used;\par
\hspace*{1em}- names containing n, count, or steps: use a positive integer or a short array.\par
4. Ensure that the arguments do not create obvious physical contradictions (e.g., the sum of partial pressures must not exceed the total pressure, and mole fractions must not sum to clearly more than 1).\par
5. Do not invent arguments that the function does not declare, and do not rename arguments.\par
6. Do not output explanatory text, comments, or extra keys; output JSON only.\par
7. JSON structure: \texttt{\{"input\_para": \{"<param1>": <value>, "<param2>": <value>, ...\}\}}, where array values are written directly as list literals (without quotes), string arguments keep their quotes, and Boolean arguments use True/False.\par
8. Avoid random numbers (do not use \texttt{np.random}).\par
9. If the function has \texttt{*args} or \texttt{**kwargs}, ignore them and handle only explicitly named arguments.\par
10. If there are dependencies between arguments (e.g., the length of one array must match another argument), satisfy the code logic.\par
11. Choose all values so that the computation does not produce numerical errors such as division by zero, the logarithm or square root of a negative number, or overflow from an overly large exponent.\par
12. If the applicable range of the formula imposes limits, choose values within that range; if it cannot be parsed, use conservative, conventional physical ranges.\par
\#\# Output\par
Output only the final JSON, with no prefix or suffix, no Markdown code block, and no additional explanation.\par
\#\# Example (illustrative, not a real function)\par
\texttt{\{"input\_para": \{"T": 298.15, "P": 101325.0, "y": [0.21, 0.79], "k0": 1.2e5\}\}}\par
(Do not include this example in the formal output.)
\end{promptblockbreakable}

\begin{promptblockbreakable}[Question Verbalization Prompt]
\#\# Identity\par
You are an expert in both Physics and Physical Chemistry, with strong proficiency in quantitative problem design and solution explanation. You rigorously apply dimensional analysis, significant figures, and domain-appropriate terminology.\par
\#\# Tasks\par
Task 1: Create a calculation problem based on the provided formula metadata and input parameters (the parameters will be injected separately as structured data; do NOT list or restate them explicitly).\par
Task 2: Provide a detailed, logically ordered solution that applies the given formula correctly, with proper unit handling and clear justification of each step.\par
Task 3: Ensure the final numerical result exactly matches the provided answer parameter.\par
\#\# Scope\par
You may generate a problem in any of the following subjects:\par
- Classical mechanics\par
- Electromagnetism\par
- Optics\par
- Thermodynamics and statistical physics\par
- Quantum mechanics\par
- Atomic, nuclear, and particle physics\par
- General physics\par
- Physical chemistry\par
- Quantum chemistry\par
provided the selected scenario is realistic and consistent with the formula's applicable range.\par
\#\# Formula Information\par
The following structured block defines the formula context:\par
\texttt{\{"name": "\{name\}", "formula": "\{formula\}", "applicable\_range": "\{applicable\_range\}"\}}\par
\#\# Reference Code (for reference)\par
\texttt{\{function\_code\}}\par
\#\# Parameter Context\par
The problem should be based on the following parameters:\par
\texttt{\{param\_list\}}\par
\#\# Expected Answer Constraint\par
The computed final result MUST numerically match:\par
\texttt{\{answer\_para\}}\par
\#\# Problem Construction Requirements\par
1. Scientific Accuracy:\par
\hspace*{1em}- Correctly apply the formula: \texttt{\{formula\}}\par
\hspace*{1em}- Maintain unit consistency; show conversions where non-trivial.\par
\hspace*{1em}- Select or invent a plausible scenario (experiment, natural process, or theoretical model) that justifies the formula's use.\par
2. Logical Clarity:\par
\hspace*{1em}- Present a coherent narrative (e.g., reaction mechanism, energy balance, motion description).\par
\hspace*{1em}- If a chemical reaction is used, ensure it is balanced and physically consistent.\par
\hspace*{1em}- If a physics scenario is used, ensure forces, fields, or energies are plausible.\par
3. Presentation Quality:\par
\hspace*{1em}- Use precise technical language.\par
\hspace*{1em}- Embed equations in LaTeX: e.g., \texttt{\$F=ma\$}, \texttt{\$q=mc\textbackslash Delta T\$}, \texttt{\$PV=nRT\$}.\par
\hspace*{1em}- Use appropriate significant figures (justify if rounding).\par
4. Do NOT:\par
\hspace*{1em}- Expose raw input parameter names.\par
\hspace*{1em}- Fully provide the formula itself that needs to be asked about.\par
\hspace*{1em}- Ask the user for additional data.\par
\hspace*{1em}- Include meta-instructions or placeholders in the final output.\par
\hspace*{1em}- Deviate from the provided answer value.\par
\#\# Solution Requirements\par
- Show step-by-step derivation.\par
- Explicitly write the main governing formula before substitution.\par
- Show unit cancellation or transformation.\par
- Justify any approximations.\par
- Conclude by enclosing the final answer (that matches the target value) in \texttt{\textbackslash boxed\{\}}.\par
- Both the generated questions and answers should be in English.\par
\#\# Output JSON Format\par
Return ONLY a JSON object with the following keys:\par
\texttt{\{"question": "A fully self-contained, well-structured Physics or Physical Chemistry problem statement (with LaTeX math expressions).", "answer": "Detailed solution showing reasoning, substitutions, unit analysis, and final matching result.", "answer\_units": "LaTeX-formatted unit expression (e.g., \textbackslash text\{mol$\cdot$L\}\textasciicircum\{-1\}, \textbackslash text\{kJ$\cdot$mol\}\textasciicircum\{-1\}, \textbackslash text\{Pa$\cdot$s\}); use 'dimensionless' if no unit."\}}\par
\#\# Additional Notes\par
- Prefer realistic constants (e.g., \texttt{\$R=8.314\textbackslash\ \textbackslash text\{J$\cdot$mol\}\textasciicircum\{-1\}\textbackslash text\{K\}\textasciicircum\{-1\}\$}) unless the scenario mandates a specific variant.\par
- If assumptions (e.g., ideal gas, negligible heat loss, non-relativistic motion) are required, state them explicitly.\par
- The style should be professional and suitable for advanced undergraduate or early graduate level.\par
\#\# Escaping requirement\par
- When showing LaTeX, JSON, file paths, or any string that normally contains backslashes, ensure each backslash is doubled (for example, write \texttt{"\textbackslash\textbackslash Delta"} for \texttt{"\textbackslash Delta"}, write \texttt{"C:\textbackslash\textbackslash path\textbackslash\textbackslash my\_file"} for \texttt{"C:\textbackslash path\textbackslash my\_file"}).\par
- Validate final output to ensure no lone \texttt{'\textbackslash'} characters remain; prefer double-escaping where ambiguity may occur.
\end{promptblockbreakable}

\begin{promptblock}[Problem Plausibility Filter Prompt (translated)]
\raggedright
[Task] Judge whether the following scientific problem is reasonable. Reasonableness is judged from the following aspects:\par
1. Whether the numerical values have practical meaning (e.g., a count cannot be a fraction, a weight cannot be negative, and the temperature range should be reasonable).\par
2. Whether units and dimensions match (e.g., a speed cannot be expressed in kilograms, and a number of molecules should not be negative).\par
3. Whether the problem conforms to basic scientific principles or common sense (e.g., an object cannot move faster than light, mass is conserved in chemical reactions, and the number of biological individuals cannot be negative).\par
[Problem]\par
\texttt{\{question\}}\par
Return the following format:\par
Conclusion: YES or NO\par
Reason: the specific reason (briefly explain why the problem is or is not reasonable).
\end{promptblock}

\begin{promptblock}[LaTeX Normalization Prompt]
\raggedright
Convert all units in the following text to proper LaTeX format with these requirements:\par
- Keep the original question structure and content unchanged.\par
- Express all units, constants, chemical formulas, and symbols with superscripts using LaTeX syntax (e.g., \texttt{\$\textbackslash text\{g/cm\}\textasciicircum 3\$}, \texttt{\$\textbackslash text\{mol/L\}\$}, \texttt{\textbackslash(\textbackslash Delta k\textbackslash)}, \texttt{\$\textbackslash text\{H\}\_2\textbackslash text\{O\}\$}).\par
- Ensure the text remains a natural and coherent question.\par
- Return only the modified question, without any explanation or extra content.\par
\medskip
Original text:\par
\texttt{\{text\}}
\end{promptblock}

\begin{promptblock}[Teacher Prompt (user message)]
\raggedright
\texttt{\{problem\} \{unit\_sentence\}}\par
Please reason step by step, and put your final answer within \texttt{\$\textbackslash boxed\{\}\$} without including units.\par
\medskip
\emph{\texttt{\{unit\_sentence\}} is drawn uniformly from ten paraphrases of ``The unit of the answer is \texttt{\{u\}}.'', where \texttt{\{u\}} is the answer unit in inline math, and is omitted when the answer is dimensionless.}
\end{promptblock}

\subsection{Benchmark Evaluation Prompts}
\label{app:prompt_evaluation}

All benchmarks are evaluated zero-shot, and no in-context examples are used.
The SciBench unit-scaling convention follows its official evaluation code,\footnote{\url{https://github.com/mandyyyyii/scibench/blob/main/eval/eval_zeroCot.py}} and the OlympiadBench prompt follows its official English open-ended template.\footnote{\url{https://github.com/OpenBMB/OlympiadBench/blob/main/inference/code/evaluators/evaluator.py}}

\begin{promptblock}[SciBench Evaluation Prompt (user message only)]
\raggedright
\texttt{\{problem\_text\}} The unit of the answer is \texttt{\{unit\}}.\par
Please reason step by step, and put your final answer within \texttt{\$\textbackslash boxed\{\}\$} without including units.\par
\medskip
\emph{Following the official SciBench code, the power-of-ten factor is removed from \texttt{\{unit\}} and handled by the scorer.}
\end{promptblock}

\begin{promptblock}[PhysReason Evaluation Prompt]
\raggedright
\textbf{System:} You are a helpful assistant.\par
\medskip
\textbf{User:}\par
Please answer this question: \texttt{\{query\}}.\par
Please wrap the final result at the end of the derivation process using \texttt{`\textbackslash boxed\{\}`} to highlight it.
\end{promptblock}

\begin{promptblock}[PhysReason Judge Prompt]
\raggedright
Your job is to look at a question, a gold target, and a predicted answer, and return a letter ``A'' or ``B'' to indicate whether the predicted answer is correct or incorrect.\par
\medskip
Question: \texttt{\{question\}}\par
\medskip
Reference Answer: \texttt{\{gold\}}\par
\medskip
Model Answer: \texttt{\{pred\}}\par
\medskip
Evaluate the model's answer based on correctness compared to the reference answer.\par
Grade the predicted answer of this new question as one of:\par
A: CORRECT\par
B: INCORRECT\par
\medskip
Just return the letters ``A'' or ``B'', with no text around it.\par
\medskip
\emph{\texttt{\{pred\}} is the final response after the reasoning block. A response is counted as correct if the judge returns A.}
\end{promptblock}

\begin{promptblock}[OlympiadBench Evaluation Prompt (official English open-ended template)]
\raggedright
\textbf{System:} You are an AI assistant. Please answer the following Physics competition questions as required.\par
\medskip
\textbf{User:}\par
The following is an open-ended problem from an International Physics competition. \texttt{\{answer\_type\_text\}}Please calculate the answer according to the given requirements and the information provided. Please use LaTeX format to represent the variables and formulas used in the solution process and results. Please end your solution with ``So the final answer is \texttt{\{final\_answer\_text\}}.'' and give the result explicitly\texttt{\{unit\_text\}}.\par
\texttt{\{context\}}\par
\texttt{\{question\}}\par
\medskip
\emph{Slots follow the official evaluation code: they state the answer type (numerical value, expression, equation, or interval), the boxed final-answer format, and, when the answer has a unit, that the unit should not be included in \texttt{\textbackslash boxed\{\}}. The context line is omitted when a problem has no context.}
\end{promptblock}

\begin{promptblock}[TheoremQA-Physics Evaluation Prompt]
\raggedright
\textbf{System:} You are a science teacher. Provide a solution to the given problem. End with the final answer inside a single LaTeX \texttt{\textbackslash boxed\{...\}}, without units or explanation inside the box. For multiple numerical answers, put the list of numbers inside that box.\par
\medskip
\textbf{User:}\par
Problem:\texttt{\{Question\}}\par
Solution:
\end{promptblock}

\end{document}